\documentclass[authoryear,preprint,review,12pt]{elsarticle}

\usepackage{amssymb}
\usepackage{amsmath}
\usepackage{graphicx}
\usepackage{float}
\usepackage{svg}
\usepackage{multirow}
\usepackage{natbib}
\makeatletter
\newenvironment{figurehere}
{\def\@captype{figure}}
{}
\makeatletter
\setcitestyle{numbers,square}
\usepackage{url}
\usepackage{hyperref}
\usepackage{lineno}
\usepackage[T1]{fontenc}
\usepackage{mathptmx}
\begin{document}
	
	\begin{frontmatter}
		
		%% Title, authors and addresses
		
		%% use the tnoteref command within \title for footnotes;
		%% use the tnotetext command for theassociated footnote;
		%% use the fnref command within \author or \address for footnotes;
		%% use the fntext command for theassociated footnote;
		%% use the corref command within \author for corresponding author footnotes;
		%% use the cortext command for theassociated footnote;
		%% use the ead command for the email address,
		%% and the form \ead[url] for the home page:
		%% \title{Title\tnoteref{label1}}
		%% \tnotetext[label1]{}
		%% \author{Name\corref{cor1}\fnref{label2}}
		%% \ead{email address}
		%% \ead[url]{home page}
		%% \fntext[label2]{}
		%% \cortext[cor1]{}
		%% \affiliation{organization={},
			%%             addressline={},
			%%             city={},
			%%             postcode={},
			%%             state={},
			%%             country={}}
		%% \fntext[label3]{}
		
		\title{Improving Underwater Semantic Segmentation with Underwater Image Quality Attention and Muti-scale Aggregation Attention}
		
		%% use optional labels to link authors explicitly to addresses:
		%% \author[label1,label2]{}
		%% \affiliation[label1]{organization={},
			%%             addressline={},
			%%             city={},
			%%             postcode={},
			%%             state={},
			%%             country={}}
		%%
		%% \affiliation[label2]{organization={},
			%%             addressline={},
			%%             city={},
			%%             postcode={},
			%%             state={},
			%%             country={}}
		
		\author[1]{Xin Zuo}
		%\ead{zuoxin@just.edu.cn}
		
		\author[1]{Jiaran Jiang}
		%\ead{1696810514@qq.com}
		
		\author[2]{Jifeng Shen\corref{cor1}}
		\ead{shenjifeng@ujs.edu.cn}
		
		\author[3]{Wankou Yang}
		%\ead{}
		
		\address[1]{School of Computer Science and Engineering, Jiangsu University of Science and Technology, Zhenjiang, 212003, China}
		\address[2]{School of Electronic and Informatics Engineering, Jiangsu University, 
			Zhenjiang, 212013, China}
		\address[3]{School of Automation, Southeast University, Nanjing, Jiangsu, 210096, China}
		
		\begin{abstract}
			%% Text of abstract
			Underwater image understanding is crucial for both submarine navigation and seabed exploration. However, the low illumination in underwater environments degrades the imaging quality, which in turn seriously deteriorates the performance of underwater semantic segmentation, particularly for outlining the object region boundaries. To tackle this issue, we present UnderWater SegFormer (UWSegFormer), a transformer-based framework for semantic segmentation of low-quality underwater images. Firstly, we propose the Underwater Image Quality Attention (UIQA) module. This module enhances the representation of high-quality semantic information in underwater image feature channels through a channel self-attention mechanism. In order to address the issue of loss of imaging details due to the underwater environment, the Multi-scale Aggregation Attention (MAA) module is proposed. This module aggregates sets of semantic features at different scales by extracting discriminative information from high-level features, thus compensating for the semantic loss of detail in underwater objects. Finally, during training, we introduce Edge Learning Loss (ELL) in order to enhance the model's learning of underwater object edges and improve the model's prediction accuracy. Experiments conducted on the SUIM and DUT-USEG (DUT) datasets have demonstrated that the proposed method has advantages in terms of segmentation completeness, boundary clarity, and subjective perceptual details when compared to SOTA methods. In addition, the proposed method achieves the highest mIoU of 82.12 and 71.41 on the SUIM and DUT datasets, respectively. Code will be available at \href{https://github.com/SAWRJJ/UWSegFormer}{https://github.com/SAWRJJ/UWSegFormer}.
		\end{abstract}
		
		%%%Graphical abstract
		%\begin{graphicalabstract}
		%\includegraphics{grabs}
		%\end{graphicalabstract}
		
		%%Research highlights
		%\begin{highlights}
		%\item Research highlight 1
		%\item Research highlight 2
		%\end{highlights}
		
		\begin{keyword}
			%% keywords here, in the form: keyword \sep keyword
			
			%% PACS codes here, in the form: \PACS code \sep code
			
			%% MSC codes here, in the form: \MSC code \sep code
			%% or \MSC[2008] code \sep code (2000 is the default)
			Underwater Image Quality Attention, Multi-scale Aggregation Attention, Underwater Semantic Segmentation, Transformer
		\end{keyword}
		
	\end{frontmatter}
	
	%\linenumbers
	
	%% main text
	\section{Introduction}
	The field of computer vision has made significant progress in semantic segmentation, which involves assigning a class label to each pixel. Underwater image segmentation is essential for underwater navigation and seabed exploration, with applications in both military and civilian contexts. 
	However, there is a significant difference between underwater imaging and normal image acquisition in the air. This is because the absorption of light by water increases exponentially with the depth of water \cite{RF46}. Besides, the scattering phenomenon occurs when the light passes through the water or particles in the water directly into the imaging system.
	Both of these factors can result in low contrast, noisy, and uneven illumination during image acquisition, which significantly reduces image segmentation performance.
	As a result, high-quality Underwater Image Enhancement plays an important role in improving the performance of underwater semantic segmentation.
	%However, compared to natural images, high-quality underwater images suitable for training deep learning models are very limited due to the low-light environment underwater that severely affects image acquisition. 
	%The lack of large datasets of high-quality underwater images is one of the main obstacles to deep learning-based underwater semantic segmentation. 
	%To reduce the need for high-quality datasets, we improve underwater semantic segmentation performance by enhancing features in low-quality underwater images.\par
	
	%One of the intuitive ideas is to utilize image enhancement techniques to improve semantic segmentation performance. 
	Recent methods for Underwater Image Enhancement (UIE) can be categorized into three groups: visual prior-based, physical model-based, and data-driven. First, prior-based methods \cite{RF1}, \cite{RF2} mainly enhance the visual performance of underwater images by adjusting pixel information, including parameters such as contrast and brightness. On the other hand, physical model-based UIE methods \cite{RF3}, \cite{RF4} can be argued to produce sharper images by correctly estimating the medium transport and other imaging parameters (e.g. uniform background light). The data-driven methods \cite{RF5}, \cite{RF6}, \cite{RF7} have shown impressive results in the Underwater Image Enhancement using deep learning techniques. Despite the significant development of all three groups of methods, their application to underwater semantic segmentation still has room for improvement. 
	Priori-based methods have insufficient understanding of underwater physical degradation processes. Physical model-based methods have modelling assumptions that are not always tenable in complex and diverse real-world underwater scenarios. %making it difficult to analyze many factors simultaneously. 
	Data-driven methods are that current Underwater Image Enhancement datasets often lack semantic segmentation annotation. 
	Therefore, this paper proposes an UIQA module that addresses the issue of uneven attenuation of underwater image channels and enhances the quality of underwater image features by enhancing the high-quality semantic information in the underwater image feature channels to achieve the effect of Underwater Image Enhancement.\par
	In addition, multi-scale feature fusion can help to capture a wider range of contextual information in visual tasks. The incorporation of multi-scale features into models can facilitate a more comprehensive understanding of the context within an image, thereby enhancing the model's ability to perceive the details of underwater targets. Inspired by classical multi-scale fusion methods \cite{RF14}, \cite{RF23}, we propose the MAA module, which aims to use the large perceptual fields of high-level features to guide the representation of low-level features, thus enabling the fusion of multi-scale features. This process facilitates the capture of more semantic information pertaining to underwater details, thereby enhancing the performance of the model.\par
	% In addition, multi-scale feature fusion can help to capture a wider range of contextual information in visual tasks. By considering features at multiple scales, models can better understand the context in an image more comprehensively and improve the perception of the target or scene. Inspired by classical multi-scale fusion methods \cite{RF14}, \cite{RF23}, we propose the MAA module, which aims to fuse multi-scale features by using the large receptive fields of high-level features to guide the representation of low-level features. This process captures more underwater semantic information, and improves the model performance. 
	In order to construct an end-to-end model for underwater image feature enhancement and semantic segmentation, we introduce the UWSegFormer, which comprises the UIQA and MAA modules within the Transformer framework. The encoder comprises of hierarchical transformers which generate both high and low-resolution feature maps to better represent the input image. These features are fed into UIQA, which enhances the high-quality semantic information in the underwater image feature channels, and then these enhanced multi-scale features are aggregated into a single scale feature by multi-scale aggregation in MAA. Furthermore, we introduce ELL during training, which provides additional information to the network by emphasising the edge information of the output mask, thereby enhancing the perception of semantic boundaries and improving the model's performance.
	% To implement Underwater Image Enhancement and multi-scale feature fusion in an unified framework, we introduce UWSegFormer, which consists of UIQA and MAA modules in the Transformer framework. The encoder comprises of hierarchical transformers which generate both high and low-resolution feature maps to better represent the input image. These features are fed into the UIQA, through which pays attention is paid to channels with severely degraded color quality in the features, and these enhanced multi-scale features are then subjected to multi-scale aggregation to single-scale features in the MAA.
	We perform underwater semantic segmentation experiments on both SUIM and DUT datasets, and the experimental results show the superiority of our method.\par
	% In order to make image enhancement and semantic segmentation in a unified framework, we introduce UWSegFormer, which features Underwater Image Quality Attention and Muti-scale Aggregation Attention in the Transformer framework. The encoder comprises of hierarchical transformers which generate both high and low-resolution feature maps to more effectively portray the input image. These features can be refined by UIQA for channels suffering from severe image quality degradation prior to being given to the decoder, but lead to increased computational cost. The MAA module based on \cite{RF14}, aims to fuse four feature maps and use the large receptive fields of the high-level features to guide the representation of the low-level features, which requires very little computation. The experimental results show the superiority of our method.\par
	Our main contributions are summarized in the following:
	\begin{enumerate}[(1)]
		% \item The Underwater Image Quality Attention is proposed based on the Transformer to achieve image feature enhancement by focusing on high quality image channels along the channel axis;
		\item The Underwater Image Quality Attention is proposed based on the Transformer, which realizes feature enhancement of underwater images by focusing high-quality image channels along the channel axis to improve the recognition ability of the model on underwater images;
		% \item The Muti-scale Aggregation Attention is proposed to efficiently use multi-scale information and fuse multiscale feature maps to focus on the details of underwater targets to improve segmentation accuracy;
		\item The Muti-scale Aggregation Attention is proposed to efficiently use different semantic information between multi-scale features, aggregating effective information of multi-scale features, so that the model focuses on the details of the underwater target and improves the segmentation accuracy;\par
		% \item The ELL is proposed to improve the perceptual range of the model by supplementing the boundary information of the learning output mask, thereby enabling the model to predict more accurately;
            \item Edge Learning Loss is designed to enhance the model’s ability to learn the target boundary by incorporating supplementary boundary information from the predicted output mask. This approach expands the model’s perceptual field, thereby improving its accuracy in boundary prediction;
		% \item The UWSegFormer is proposed to unify the underwater image feature enhancement and semantic segmentation model in a framework for semantic segmentation\textbf{} task in complex underwater environment.
            \item The UWSegFormer is proposed to unify the underwater image feature enhancement and semantic segmentation model in a framework for semantic segmentation\textbf{} task in complex underwater environment;
            \item We have conducted extensive experiments on both the SUIM and DUT datasets to demonstrate that our method outperforms existing SOTA methods under underwater conditions.
	\end{enumerate}\par
    The organization of the rest paper is shown below. Section 2 presents the related work. Section 3 describes our proposed algorithm. Section 4 provides all experiments and analysis. Section 5 summarizes the entire paper and outlines future research directions.
	
	\section{Related Work}
    In this section, the current task of underwater semantic segmentation focuses on three main areas:Underwater Image Enhancement, Multi-scale feature fusion and Transformer with semantic segmentation.
	\label{}
	\subsection{Underwater Image Enhancement}
	Water is a highly scattering medium, which reduces the amount of light reaching the camera by scattering it in multiple directions. As a result, underwater images are often blurred, murky and monochromatic. To address these issues and reduce the reliance on manual labeling of real-world underwater datasets, Li et al. \cite{RF15} proposed an unsupervised approach for WaterGAN to generate images like underwater scenes using depth maps and aerial images, and the generated images are used for further training. Furthermore, \cite{RF16} demonstrated a weakly-supervised color correction underwater model based on CycleGAN \cite{RF17}, where \cite{RF16} directly used the network structure of CycleGAN \cite{RF17}, while the use of adversarial networks and the construction of multiple loss functions allowed the network to use unpaired underwater images during training, which greatly improving the generalizability of the network model to complex underwater environments. Recently, Li et al. \cite{RF18} introduced WaterNet which is a gated fusion network and \cite{RF18} enhances underwater images by using predicted confidence maps as inputs leading to better results. Yang et al. \cite{RF19} introduced a Conditional Generative Adversarial Network to generate the clear underwater images by constraining the input conditions. 
    % However, due to the limitations of the GAN-based models, the above models generate unstable enhancement results, which is not conducive to the stability of training.\par
	Furthermore, Ucolor \cite{RF20} is an Underwater Image Enhancement network that incorporates physical model-based media transmission control. This guidance enhances the network's focus on regions of degraded quality and improves the visual quality of the network's output. \par
 %    Nevertheless, the physical model may not always provide accurate results as the underwater environment changes.\par
	% In this paper, we consider the problem of uneven attenuation of underwater image channels and investigate methods for enhancing the feature representation of underwater images without introducing additional inputs.
	\subsection{Multi-scale feature fusion}
	The top layer feature maps are usually rich in semantic information but have low resolution. On the contrary, the lower layer feature maps are the exact opposite of the top layer feature maps. To solve this problem, \cite{RF21}, \cite{RF22} directly merged feature maps of different layers. However, due to the semantically weak low-level features generated by traditional classification networks, making these multi-scale features are unsuitable for downstream intensive prediction tasks. \par
	To address this problem, for semantic segmentation, DeeplabV3+ \cite{RF23} enhanced the spatial pyramid pooling (SPP) module to detect multi-scale convolutional features by applying null convolution with image-level features at different expansion coefficients, and fused low-level features with semantically strong high-level features obtained by SPP to improve segmentation boundary accuracy. Eseg \cite{RF24} used a richer multi-scale feature space and incorporated a powerful feature fusion network for effective feature integration. \par
	% However, the previously mentioned methods don’t consider the effect of underwater low illumination environment on the details of underwater objects, and don’t notice the semantic connection between high-level and low-level features during the fusion process. This results in the loss of detailed semantic information of the fused features, which significantly reduces the accuracy of the segmentation process. Therefore, we aim to explore the possibility of using the large receptive field of high-level features to guide the performance of low-level features and to use the connection between multi-level features for multi-scale feature aggregation to capture more local detail information.
	\subsection{Transformer with semantic segmentation}
	Semantic segmentation extends image classification to the task of identifying objects at the pixel level. Deep learning methods have revolutionised the field, making semantic segmentation much more accurate and efficient. Fully Convolutional Networks (FCN) \cite{RF25} have enabled end-to-end prediction at the pixel level without the need for manual feature extraction. Based on FCN, in studies of marine life, extensive researches have focused extensively on semantic segmentation using Convolutional Neural Networks (CNN) \cite{RF26}. The WaterSNet \cite{RF27} represented a modified CNN model that incorporated an attention mechanism. The attentional fusion block was used to exploit global contextual information. Additionally, a receptive field block has been used to extract multi-scale features in the WaterSNet. USS-Net \cite{RF49} employs an auxiliary feature extraction network to enhance the backbone network's ability to capture semantic information and incorporates a channel attention mechanism to highlight key target regions. Additionally, it utilizes multi-stage feature input up-sampling, which preserves more boundary details when restoring high-resolution features. Furthermore, the model integrates cross-entropy loss and Dice loss to improve the precise recognition of target regions. DeepLab-FusionNet\cite{RF47} improves DeepLabV3+ by incorporating a multi-resolution parallel branching structure to improve multi-scale feature extraction. Additionally, it employs an optimized inverted residual structure as the fundamental feature extraction module in the encoder, thereby enhancing feature representation and segmentation accuracy. However, these underwater models still rely on CNNs, which limits their ability to make long-range connections across the image and extract only local information. \par
	Recent approaches have demonstrated the achievements of the Transformer structure in the field of semantic segmentation \cite{RF10}, \cite{RF28}. The first model to effectively apply Transformer to the vision domain was ViT. When an image was fed into the ViT, it was segmented into a series of patches, which were distributed and passed as an input sequence to the Transformer model. Following ViT, several enhanced approaches have been developed. In terms of training, DeiT \cite{RF29} performed well in efficient training with limited datasets and showed better generalization performance. To address multi-scale processing, Twins \cite{RF12} explored into the fusion of local and global self-attention mechanisms. Another advance was CrossViT \cite{RF30}, which introduced a dual-path transformer architecture specifically designed to handle tokens of different scales. For hierarchical design, Swin Transformer \cite{RF11}, PVT \cite{RF31} and LVT \cite{RF32} used a four-stage design with gradual down-sampling of feature maps, facilitating downstream tasks.In the field of underwater image segmentation, Chen et al. \cite{RF48} introduced an improved SegFormer by integrating an Efficient Multiscale Attention (EMA) mechanism into the encoder to enhance multi-scale feature extraction. Furthermore, they incorporated a Feature Pyramid Network (FPN) structure into the decoder to fuse feature maps at multiple resolutions, enabling the model to effectively integrate contextual information and enhance segmentation performance. \par
    Nevertheless, the physical models and the GAN models may not always provide accurate results as the underwater environment changes. Meanwhile, the previously mentioned methods don’t consider the effect of underwater low illumination environment on the details of underwater objects, and don’t notice the semantic connection between high-level and low-level features during the fusion process. This results in the loss of detailed semantic information of the fused features, which significantly reduces the accuracy of the segmentation process. None of the above mentioned methods, although showing a high degree of effectiveness, take into account the fact that aqueous environments are more likely to cause a loss of image colour.\par
In summary, in this paper, we consider the problem of non-uniform attenuation of underwater image channels and investigate methods to enhance underwater image feature representation without introducing additional inputs. At the same time, we are exploring the possibility of using the large sensory fields of high-level features to guide the representation of low-level features, and utilizing the linkage between multi-level features for multi-scale feature aggregation to capture more local detail information. Finally, the model's recognition of target boundaries is improved by adding an edge loss function to the traditional semantic segmentation model.
	% While the above Transformer methods have demonstrated high effectiveness in non-underwater experiments, their performance falls short when applied to underwater datasets. This is due to the complex water environment, which is more likely to cause color loss in images compared to the common air medium on land. In this paper, we aim to improve the performance of the model by adding an underwater image feature enhancement module and a new multi-scale fusion module to the traditional semantic segmentation model.
	\begin{figure}[H]
		\centering
		\includegraphics[width=1\linewidth]{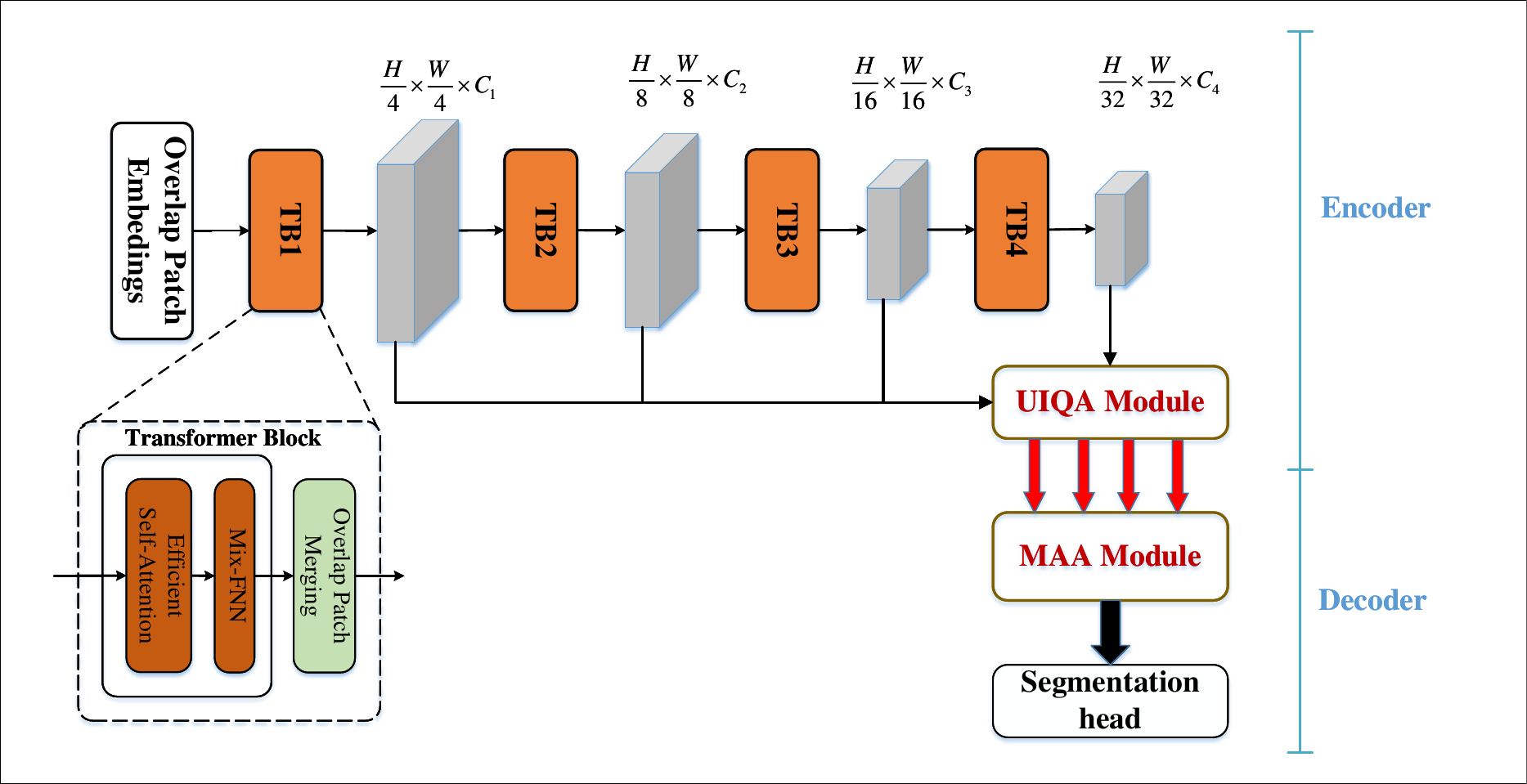}
		\caption{Framework of the proposed UWSegFormer. The encoder part with hierarchical Transformer and UIQA to extract coarse and fine features. The decoder part with MAA to exploit multi-level features and predict semantic segmentation masks by aggregation. During the training, the ELL is incorporated into the loss function.}
		\label{fig:uwsegformer}
	\end{figure}

	\section{Method}
	We propose the UWSegFormer model, for the underwater semantic segmentation task. In Section \textbf{3.1}, we first revisit the comprehensive framework of segformer model as illustrated in Fig.~\ref{fig:uwsegformer}.
	Our method improves SegFormer with novel components: \textbf{(1)} Underwater Image Quality Attention, whose main role is to collect the feature maps extracted by the previous Transformer module and re-encode them to enhances the quality of underwater image features by enhancing the high-quality semantic information in the underwater image feature channels (Section \textbf{3.2}). \textbf{(2)} Muti-scale Aggregation Attention, which efficiently fuses the feature maps provided by the encoder to improve segmentation accuracy (Section \textbf{3.3}). \textbf{(3)} Edge Learning Loss, whitch learns boundary additional information from output mask to improve the accuracy of the model's predictions (Section \textbf{3.4}).

	\subsection{UWSegFormer Architecture}
	In this section, we first provide a brief summary of the standard SegFormer, which consists of two modules: a hierarchical transformer encoder and a lightweight ALL-MLP decoder \cite{RF13}. The encoder generates multi-scale feature maps with different resolutions. These multi-scale features are then fed to the decoder for feature fusion to generate the final mask. \par
	The encoder in SegFormer consists of four stages that generate feature maps $ {F}_1, {F}_2, {F}_3, {F}_4$ with resolution of $1/4$, $1/8$, $1/16$ and $1/32$. Each stage includes overlapped patch merging (OPM), efficient self-attention (ESA) and Mix-FNN (MF) modules, which are shown in the TransFormer Block of Fig.~\ref{fig:uwsegformer}. \par
	Then, SegFormer passes the feature maps $F_i$ in stage $i$ of the encoder to the ALL-MLP decoder to predict the segmentation mask $M$ with size of $\frac{H}{4}\times\frac{W}{4}\times N_{cls}$, where $N_{cls}$ denotes the number of categories. The decoder can be formulated in Eq. (\ref{eq4})-(\ref{eq7}). 
	\begin{equation}\label{eq4}
		{{F}_i=Linear\left(C_i, C\right)\left(F_i\right)}
	\end{equation}
	\begin{equation}\label{eq5}
		{{F}_i=Up\left(\frac{H}{4}\times\frac{W}{4}\right)\left(F_i\right)}
	\end{equation}
	\begin{equation}\label{eq6}
		{{F}_i=Linear\left(4C,C\right)\left(Concat\left({F}_i\right)\right)}
	\end{equation}
	\begin{equation}\label{eq7}
		M=Linear\left(C,N_{cls}\right)\left(F\right)
	\end{equation}
	where $i\in\left\lbrace1,2,3,4\right\rbrace$. $Linear({C}_{in}, {C}_{out})(\cdot)$ refers to a linear layer with ${C}_{in}$ and ${C}_{out}$ as the dimensions of the input and output vectors, respectively. ${C}_{i}$ denotes the dimension of the ${i}^{th}$ layer features, and $C$ is the set embedding dimension. Through $Up(\cdot)$ features are up-sampled to $\frac{H}{4}\times\frac{W}{4}$ and then concatenated by $Concat(\cdot)$.\par
	Fig.~\ref{fig:uwsegformer} shows the overall architecture of UWSegFormer. Similar to existing SegFormer-based approaches such as MTLSegFormer \cite{RF33}, our method also uses a classical encoder-decoder structure with standard four-stage design. The main difference between SegFormer and our proposed method is that we add UIQA to SegFormer’s Transformer encoder and improve the conventional ALL-MLP decoder with MAA module. The UIQA is designed to identify and analyse image feature channels that contain high-quality semantic information, whereas the MAA is designed to make more effective use of multi-scale information in the decoder. During training, we add ELL to the loss to improve the model's perception of the boundary information. The following three subsections describe the detailed structure of the UIQA, MAA and ELL.
	\begin{figure}[H]
		\centering
		\includegraphics[width=1\linewidth]{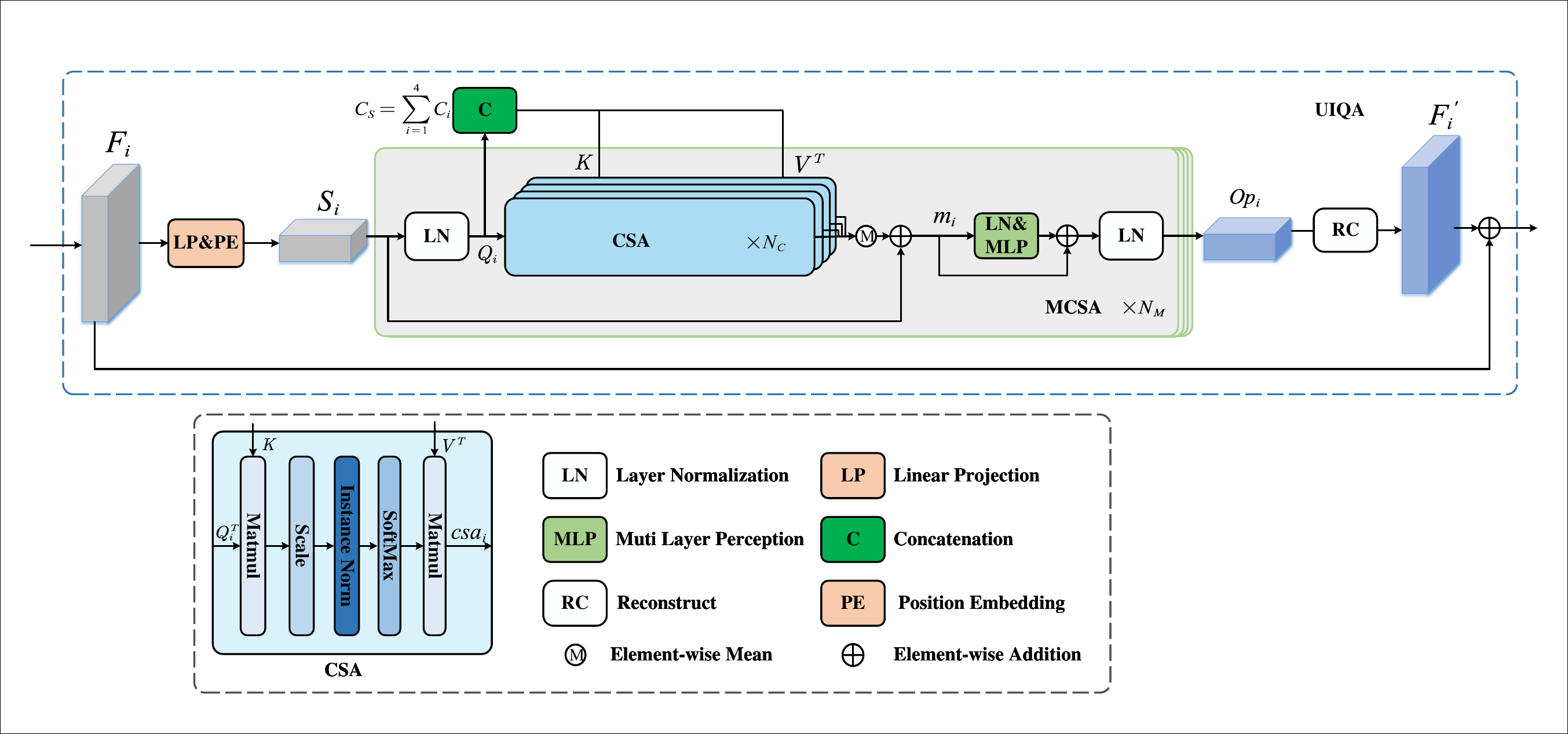}
		\caption{Detailed structure of the UIQA module.}
		\label{fig:cda}
	\end{figure}
	\subsection{Underwater Image Quality Attention}
	In order to pay attention to underwater image feature channels with high-quality semantic information, we are inspired by \cite{RF34}, \cite{RF35} and design the UIQA to be plugged into the encoder (in Fig.~\ref{fig:cda}), which consists of three components: Feature Encoding, Multi-scale Channel Self-Attention and Reconstruct blocks.\par
	\textbf{Feature Encoding.}  First, the inputs to UIQA are feature maps $
	{F}_{i}\in\\
	{\mathbb{R}}^{\frac{H}{{2}^{i+1}}\times\frac{W}{{2}^{i+1}}\times{C}_i}\left(i=1,2,3,4\right)
	$. Convolution kernel with filter size of $\frac{P}{{2}^{i}}\times\frac{P}{{2}^{i}}\left(i=1,2,3,4\right)$ and step size $\frac{P}{{2}^{i}}\left(i=1,2,3,4\right)$ are used to perform a linear projection on the feature maps. After that, we obtain four feature maps $S_i\in\mathbb{R}^{d\times C_i}\left(i=1,2,3,4\right)$, where $d=\frac{HW}{{P}^{2}}$. Next, to divide the feature maps into an equal division of consistent blocks and maintain the same number of channels $C_i (i=1,2,3,4)$, $P$ is set to 32. Then, $K\in{\mathbb{R}}^{d\times C}$, ${Q}_i\in{\mathbb{R}}^{d\times{C}_i}\left(i=1,2,3,4\right)$ and $V\in{\mathbb{R}}^{d\times C}$ can be obtained from Eq. (\ref{eq8}).
	
	\begin{equation}\label{eq8}
		K=SW_K,Q_i=S_iW_{Q_i},V=SW_V
	\end{equation}
	where $W_K\in{\mathbb{R}}^{d\times C}$,$W_{Q_i}\in{\mathbb{R}}^{d\times{C}_i}\left(i=1,2,3,4\right)$and $W_V\in{\mathbb{R}}^{d\times C}$ are the learnable weight matrices; $S$ is generated by the channel dimensionality connection ${S}_i\in{\mathbb{R}}^{d\times{C}_i}\left(i=1,2,3,4\right)$, where $C_{S}=C_1+C_2+C_3+C_4$. In this work, $C_1$, $C_2$, $C_3$ and $C_4$ are set to be 32, 64, 160, 256 respectively.\par
	\textbf{Multi-scale Channel Self-Attention (MCSA).} The Channel Self-Attention block inputs $K\in{\mathbb{R}}^{d\times C}$, $V\in{\mathbb{R}}^{d\times C}$ and ${Q}_i\in{\mathbb{R}}^{d\times{C}_i}\left(i=1,2,3,4\right)$, and outputs the ${csa}_i\in{\mathbb{R}}^{{C}_i\times d}\left(i=1,2,3,4\right)$ which is formulated in Eq. (\ref{eq9}),
	\begin{equation}\label{eq9}
		csa_{i}=S o f t m a x\left(I N\left({\frac{Q_{i}^{\ T}K}{\sqrt{C_{S}}}}\right)\right)V^{T}
	\end{equation}
	where $IN(\cdot)$ denotes instance normalization operation and all $Q_i$ share $K$ and $V$.\par 
	While traditional methods use spatial dimensional self-attention mechanisms, MCSA uses a channel self-attention mechanism along the channel axis rather than the patch-axis. The advantage of our model evaluates the entire image channel to capture the overall quality issues so that the network can pay attention to the underwater image feature channels which contain high-quality semantic information. The $j$th MCSA layer output can be expressed in Eq. (\ref{eq10}), 
	\begin{equation}\label{eq10}
		m_i={\frac{csa_{i}^1+csa_{i}^2+...+csa_{i}^{N_C}}{N_C}}+Q_i
	\end{equation}
	where $N_C$ is the number of heads.
	Just like the forward propagation in [9], the expression for the Feedforward network (FFN) output ${Op}_i$ in Eq. (\ref{eq11}),
	\begin{equation}\label{eq11}
		Op_{i}=LN\bigl(m_{i}+M L P(L N(m_{i}))\bigr)
	\end{equation}
	where ${Op}_i\in{\mathbb{R}}^{d\times{C}_i}\left(i=1,2,3,4\right)$, $MLP(\cdot)$ means multi-layer perception. $LN(\cdot)$ represents layer normalization.\par
	\textbf{Reconstruct.} Finally, the sequences of output features ${Op}_i\in{\mathbb{R}}^{d\times{C}_i}\left(i=1,2,3,4\right)$ are reconstructed to generate feature maps $F^{'}_{i}\in{\mathbb{R}}^{\frac{H}{{2}^{i+1}}\times\frac{W}{{2}^{i+1}}\times{C}_i}\left(i=1,2,3,4\right)$, which has the same size with the input features $F_{i}$. The output of the $i$th feature is formulated in Eq. (\ref{eq12}).
	\begin{equation}\label{eq12}
		F^{'}_{i}=Relu(Norm(Conv(Up(Op_{i}))))+F_{i}
	\end{equation}
	where $Up(\cdot)$ denotes up-sampling function. These features are provided to the convolutional layer to produce a reconstructed feature map by a weighted combination of the input features. Next, batch normalization and an activation function ($Relu(\cdot)$) are implemented on the generated feature maps to normalize them and introduce nonlinearity. 
	
	\subsection{Muti-scale Aggregation Attention}
	The MAA is designed to capture detail features that are attenuated by the expression of the underwater environment. We use the larger receptive fields of high-level features to direct low-level features in extracting important information to improve multi-scale feature representations. \par
\begin{figure}[H]
		\centering	\includegraphics[width=1\linewidth]{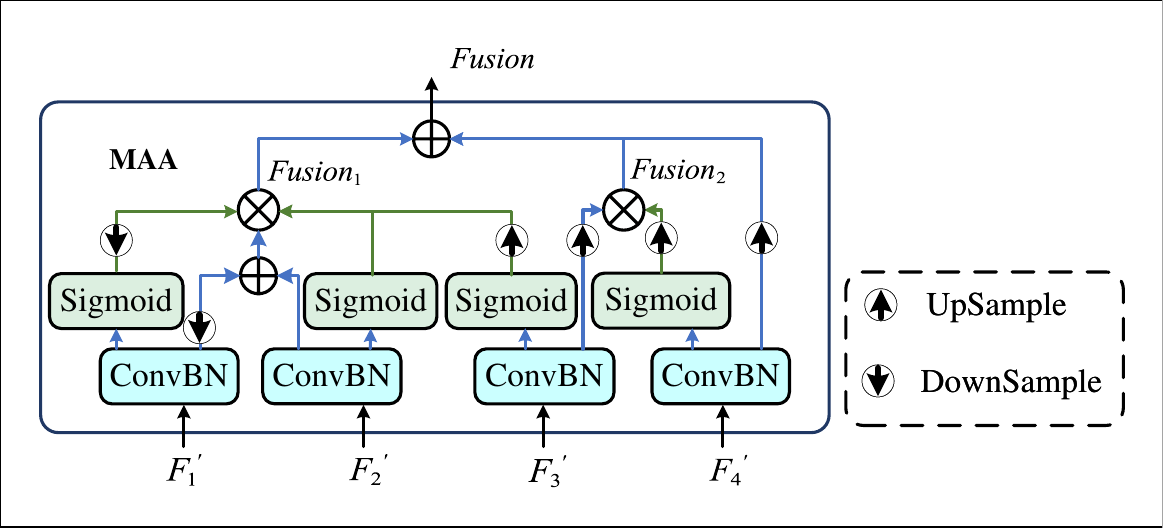}
		\caption{Detailed structure of the MAA module, where the role of the ConvBN layer is to change the channel $C_i$ of the input features into $C$.}
		\label{fig:MAA}
	\end{figure}

	 %By extracting feature maps of different scales using various extraction strategies, we can significantly reduce computational load, resulting in faster processing and lower memory usage.\par
	As shown in Fig.~\ref{fig:MAA}, with four feature maps ${F^{'}_i}\left(i=1,2,3,4\right)$ from the encoder as inputs. Among the generated feature maps, $\left\lbrace {F^{'}_3},{F^{'}_4}\right\rbrace$ have a larger receptive field and contain richer semantic information. So they act as information filters to find out the important information in the low-level features $\left\lbrace {F^{'}_1},{F^{'}_2}\right\rbrace$. In the filtering process, $\left\lbrace {F^{'}_1}  ,{F^{'}_3},{F^{'}_4}\right\rbrace$ are sampled to the same resolution as ${F^{'}_2}$, and the weight coefficients are obtained by using the sigmoid function. After the sigmoid function,$\left\lbrace {F^{'}_1}, {F^{'}_2}, {F^{'}_3}\right\rbrace$ are multiplied to filter the sum of $\left\lbrace {F^{'}_1},{F^{'}_2}\right\rbrace$ as shown in Eq. (\ref{eq13}). Meanwhile, ${F^{'}_4}$ is used to extract the semantic information in ${F^{'}_3}$ as shown in Eq. (\ref{eq14}).
	\begin{equation}\label{eq13}
		\begin{split}
			F u s i o n_{1}=Up\bigl(S C({F^{'}_3})\bigr)\times S C\bigl({F^{'}_2}\bigr)\times Down\bigl(S
			C({F^{'}_1})\bigr)\\
			\times\bigl(C o n v\bigl({F^{'}_2}\bigr)+Down\bigl(C o n v({F^{'}_1})\bigr)\bigr)
		\end{split}
	\end{equation}
	\begin{equation}\label{eq14}
		F u s i o n_{2}=Up\bigl(S C({F^{'}_4})\bigr)\times Up\bigl(C o n v({F^{'}_3})\bigr)
	\end{equation}
	where $SC(\cdot)$ is denoted as $Sigmoid(Conv(\cdot))$, $Up(\cdot)$ denotes the up-sampling function, $Down(\cdot)$ denotes the down-sampling function.\par
	Furthermore, our observation indicates that features with rich semantic information can complement the previously filtered detail features and are essential to improve model performance. Therefore, we add the feature ${F^{'}_4}$, the maximum receptive field, to the filtered detail features and use it to improve the global view. The overall process is as in Eq. (\ref{eq15}). \par
	\begin{equation}\label{eq15}
		F u s i o n=F u s i o n_{1}+F u s i o n_{2}+Up\bigl(C o n v{(}{F^{'}_4}{)\bigr)}
	\end{equation}\par
	After feature fusion, the fused feature maps $F u s i o n$ capture rich spatial and semantic information, forming the basis for segmentation performance.A dropout layer and a linear layer are then implemented to generate the final prediction mask $M$ in Segmentation head. This process helps to exchange of information across channel dimensions, as described in Eq. (\ref{eq16}).
	\begin{equation}\label{eq16}
		{M}={L}inear\bigl(C,N_{cls}\bigr)\Bigl(Dropout\bigl(Fusion\bigr)\Bigr)
	\end{equation}
	where $N_{cls}$  is the number of categories.
	\subsection{Edge Learning Loss}
Binary cross-entropy (BCE) is typically employed in semantic segmentation to compute the mask loss. However, due to the phenomenon of quality degradation, the boundaries of underwater targets are often blurry, and since the boundary pixels are considerably smaller than the mask pixels. This leads to the fact that BCE can only localise to the blurry boundary region, which significantly impedes the network's ability to learn about the boundary information. In order to extract highly accurate boundary information in the output mask, we designed two $3\times3$ convolutions with step size of 1, padding of 1, and convolution kernel of the Scharr operator. These convolutions were used to find the gradient maps in the x and y directions, respectively. The convolution kernels are shown in Eq. (\ref{eq17}).
 \begin{equation}\label{eq17}
		{k}_x=\left[
\begin{array}{llllllllll}
	-3 & 0 & 3 \\
	-10 & 0 & 10 \\
	-3 & 0 & 3
\end{array}\right]
,
{k}_y=\left[
\begin{array}{llllllllll}
	-3 & -10 & -3 \\
	0  & 0  & 0  \\
	3 & 10 & 3
\end{array}\right],
	\end{equation}
where $k_x$ denotes horizontally oriented convolution kernel and $k_y$ denotes vertically oriented convolution kernel.\par
Subsequently, the complete edge information is determined by taking the square root. As shown in Eq. (\ref{eq18}).
 	\begin{equation}\label{eq18}
{E}=\sqrt{{{conv}_x\left({M}\right)}^{2}+{{conv}_y\left({M}\right)}^{2}}
	\end{equation}
where ${E}$ denotes the complete edge information and ${M}$ denotes output mask. ${conv}_x$ denotes a 3 × 3 convolution using the $k_x$
convolution kernel and ${conv}_y$ denotes a 3 × 3 convolution using the $k_y$ convkolution kernel.\par
Finally, the total loss is shown as in Eq. (\ref{eq19}).
 	\begin{equation}\label{eq19}
L={\lambda}{_1}ELL(E,G_e)+{\lambda}{_2}BCE({M},G)
	\end{equation}
where $G$ denotes the ground truth to $M$ and $G_e$ denotes the boundary ground truth obtained by the morphological operation algorithm from mask $G$. We set $\left\{\lambda_{1},\lambda_{2}\right\}$ to $\left\{1.0,3.0\right\}$ in experiments.

    \section{Experiments}
    In this section, we mainly use two datasets, SUIM and DUT, to validate the effectiveness of the proposed module through a large number of ablation experiments, and perform data and visualization analysis.
	\subsection{Experimental Settings}
	\noindent \textbf{SUIM Dataset.} SUIM \cite{RF36} is used for semantic segmentation of natural underwater images. A total of 1525 images are used in SUIM for the training set, and 110 images are used for validation and testing. The SUIM is divided into six categories, namely Background (waterbody) BW, Human divers HD, ROVs $\&$ instruments RO, Wrecks and ruins WR, Reefs $\&$ invertebrates RI, and Fish $\&$ vertebrates FV.\par
\begin{figure}[H]
		\centering	\includegraphics[width=1\linewidth]{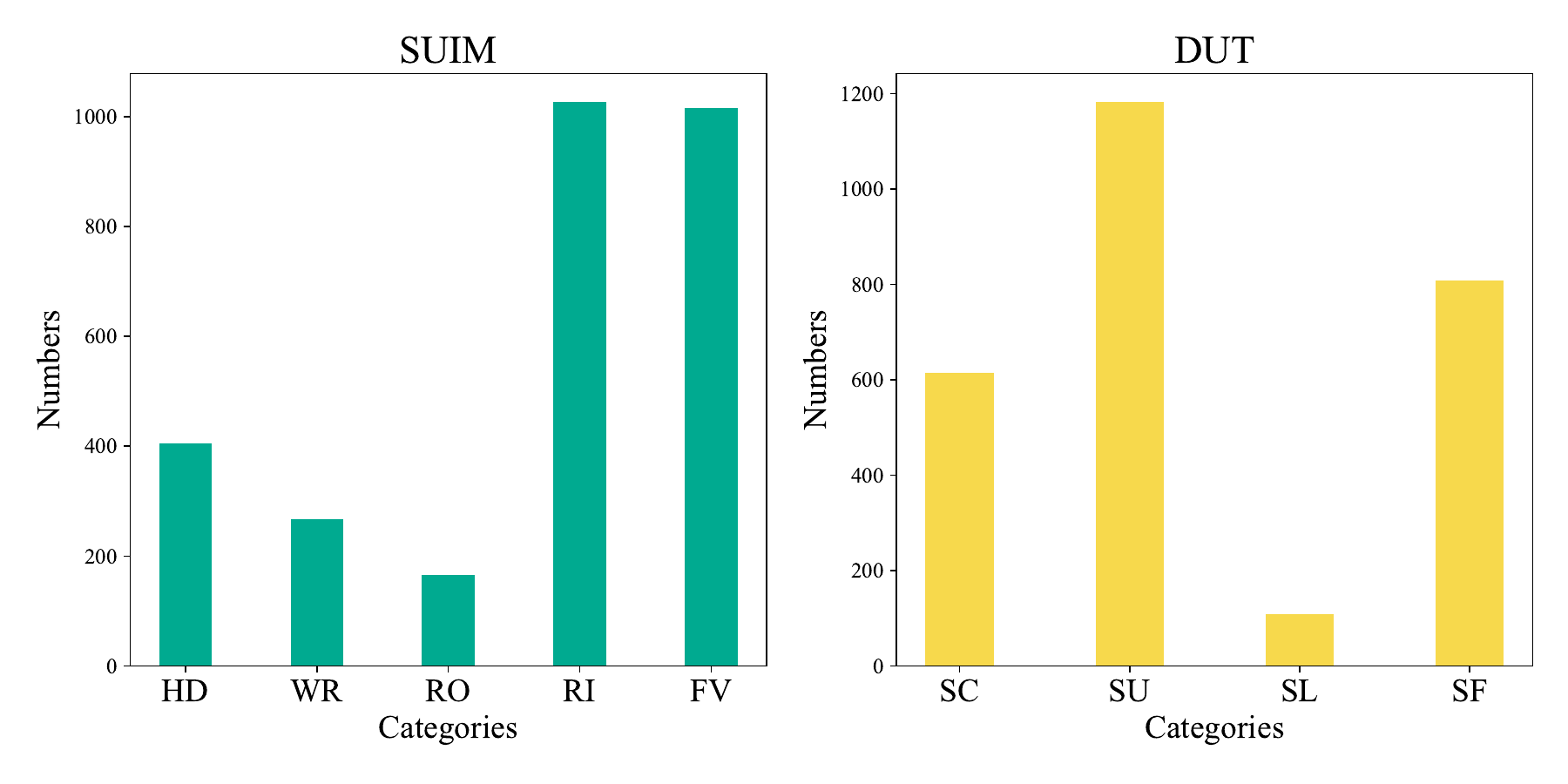}
		\caption{Details of SUIM and DUT datasets.}
		\label{fig:comparison_chart}
	\end{figure}

	\noindent \textbf{DUT Dataset.} DUT \cite{RF37} is a challenging underwater dataset in which the images contain some impurity interference. The DUT dataset consists of 6617 images, of which 1380 images with semantic segmentation annotations are selected for training and the remaining 107 images with annotations are used for validation and testing. Its categories are Background (waterbody) BW, Sea Cucumber SC, Sea Urchin SU, Scallop SL, Starfish SF. Fig.~\ref{fig:comparison_chart}. specifically shows the number of each category in both datasets.\par
	\noindent \textbf{Evaluation metrics.} Following SegFormer, we also use the Mean Intersection-over-Union (mIoU) metric on both the SUIM and DUT validation sets. This standardized evaluation allows for a comprehensive comparison of the results.\par
	\noindent \textbf{Implementation details.} We initialize the encoder with pre-trained weights on the Imagenet-1K \cite{RF38} and initialize the decoder with random values. Following SegFormer, we train our method with an initial learning rate of 0.000006 and use the AdamW \cite{RF39} optimizer with a batch size of 8 with 160K iterations when training on the SUIM and 40K iterations when training on the DUT, updated by a Poly-LR scheduling with a factor of 1. Our method is implemented in Python using the MMSegmentation \cite{RF40} codebase. The experiments are performed on a computer with a 12th Gen Intel(R) Core (TM) i9-12900k CPU, 64 GB of RAM, and an NVIDIA 3080Ti GPU, that includes 12 GB of graphics memory. \par
	For both two datasets, we use the same crop size of 640×480 while using the same data enhancement strategies including random scale ranges in [0.5, 2.0], and random horizontal flipping. We present results on a single scale on the validation set for comparison with other methods.
	
	\subsection{Ablation Studies}
	\noindent \textbf{Influence of the components.} In order to demonstrate the effectiveness of the UIQA, MAA and ELL, we performed ablation studies on the SUIM dataset. The baseline model is SegFormer. Tab.~\ref{table:1} demonstrates that using the UIQA module enables the model to pay attention those high-quality channels, resulting in enhanced performance. 
%To reduce the computational complexity of the model, 
The model using the MAA module not only markedly reduces the number of input parameters, but also demonstrates enhanced performance compared to the model employing the ALL-MLP. The introduction of ELL enhances the learning of boundary information by the model, resulting in a notable improvement in performance. Finally, our approach achieves the highest mIoU, indicating that the integration of the UIQA module, MAA module and ELL can further improve the overall performance.\par
	\begin{table}[H]
		\begin{center}
			\caption{Ablation study on the efficacy of different components in UWSegFormer.}
			\label{table:1}
			\begin{tabular}{c|c|c|c|c} % <-- Alignments: 1st column left, 2nd middle and 3rd right, with vertical lines in between
				\hline
				Models & Params(M)↓ & GFlops↓ & mIoU↑ & FPS↑\\
				\hline
				baseline & 3.72 & 7.94 & 80.57 & 95.97\\
				\hline
				w/ UIQA & 22.02 & 8.31 & 81.31 & 58.18\\
				w/ MAA & \textbf{3.47} & \textbf{2.84} & 81.04 & \textbf{121.21}\\
				w/ ELL & 3.72 & 7.94 & 81.48 & 95.97\\
				UWSegFormer & 21.78 & 3.21 & \textbf{82.12} & 62.42\\
				\hline
			\end{tabular}
		\end{center}
		
	\end{table}\par
	\noindent \textbf{Influence of the MAA channel dimension C.}  In Tab.~\ref{table:2}, we have studied the effect of the channel dimension $C$ for the MAA module on the DUT dataset. By comparing these results, we can clearly observe that when we set the channel dimension to $C=128$, the MAA module shows a decent performance, while gaining the lowest computational complexity.\par
	As $C$ increases, the performance improvement is marginal, but it is accompanied by a huge number of parameters as well as significant increase in computational latency. Therefore, we set the channel dimension to $C=128$ in the paper.\par
	
	\begin{table}[H]
		
		\begin{center}
			\caption{The effect of the MAA dimension $C$.}
			\label{table:2}
			\begin{tabular}{c|c|c|c} % <-- Alignments: 1st column left, 2nd middle and 3rd right, with vertical lines in between
				\hline
				C & GFlops↓  & FPS↑ & mIoU↑\\
				\hline
				128 & \textbf{3.21} & \textbf{62.25} & 71.41 \\
				256 & 3.65 & 54,75 & 71.42 \\
				512 & 4.99 & 53.82 & \textbf{71.49}\\
				\hline
			\end{tabular}
		\end{center}
		
	\end{table}\par
\begin{figure}[H]
		\centering	\includegraphics[width=.8\linewidth]{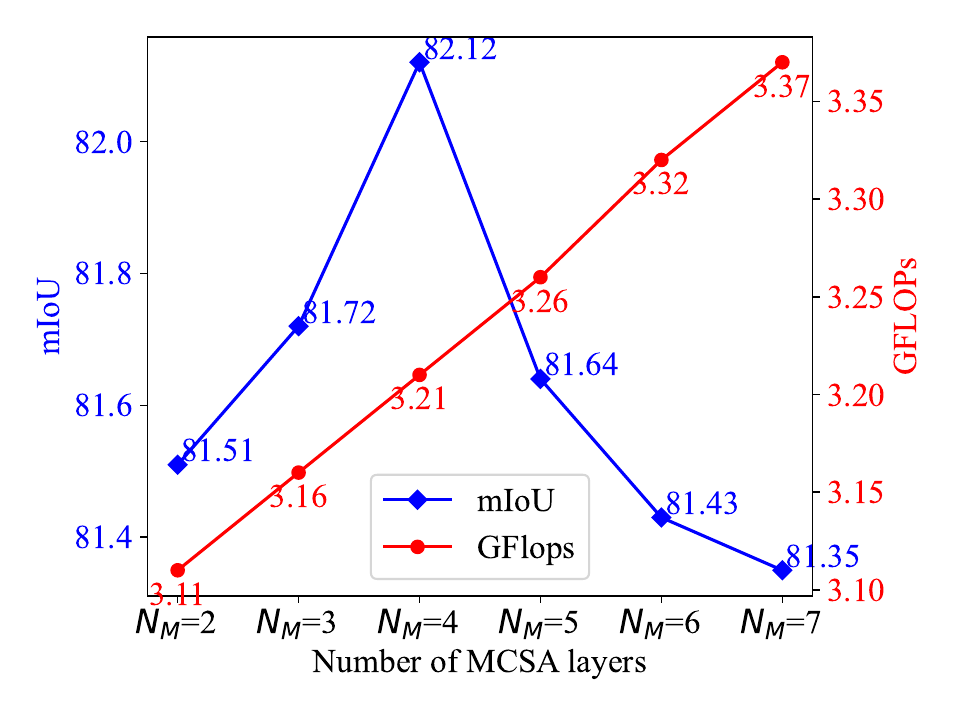}
		\caption{The effect of $N_M$ on the performance of the model. The blue line shows mIoU and the red line shows GFlops}
		\label{fig:line_chart}
	\end{figure}

	\noindent \textbf{Influence of the Numbers of MCSA layers.} In this section, we have discussed the effect of using different numbers of MCSA layers ($N_M$) on the SUIM dataset. Fig.~\ref{fig:line_chart} shows that the  computational complexity increases linearly with $N_M$. However, we have observed that the performance improvement is not significant when $N_M \le 4$, while overfitting occurs when $N_M > 4 $, which can lead to significant performance degradation. Therefore, we use $N_M=4$ as the default setting in the following experiments.
	\noindent \textbf{Influence of the Numbers of CSA heads.}The impact of the number of CSA heads ($N_C$) on accuracy was examined in Tab.~\ref{table:3}. It has been observed that the accuracy increases as $N_C$ increases when $N_C \le 4$, while performance degradation happened when $N_C \textgreater 4$. It is speculated that this is due to the ineffectiveness of attention. As $N_C$ increases, using too many attention heads can cause the model to overly focus on local details, ignoring global information. This can lead to ineffective attention and negatively impact overall performance. So, we set $N_C = 4$ as the default setting in this paper.\par
	%\subsection{UIQA Generalizability Study}
	%1.
	%\setcounter{footnote}{0}
	%\renewcommand{\thefootnote}{†}
	%%%\fnsymbol{footnote}
	
	\begin{table}[H]
		\begin{center}
			\caption{Influence of the Numbers of CSA heads ($N_C$). Experimental results of $N_C$ varying from 1 to 5.}
			\label{table:3}
			\begin{tabular}{c|c|c|c} % <-- Alignments: 1st column left, 2nd middle and 3rd right, with vertical lines in between
				\hline
				$N_C$ & Params(M)↓  & GFlops↓ & mIoU↑\\
				\hline
				1 & \textbf{14.33} & \textbf{3.10} & 81.13 \\
				2 & 16.81 & 3.14 & 81.25 \\
				3 & 19.30 & 3.17 & 81.48\\
				4 & 21.78 & 3.21 & \textbf{82.12}\\
				5 & 24.26 & 3.25 & 81.78\\
				\hline
			\end{tabular}
		\end{center}
	\end{table}\par

	\noindent \textbf{Influence of the edge detection operator.} The experimental results are presented in Tab.~\ref{table:edo}. To obtain the edge information in the output mask, several different edge detection operators were employed. The results indicate that Scharr operator achieved the best result. This is evidenced by the fact that Scharr operator is able to identify more accurate boundary locations, which demonstrates the significance of boundary information in underwater semantic segmentation.\par
%\subsection{UIQA Generalizability Study}
%1.
%\setcounter{footnote}{0}
%\renewcommand{\thefootnote}{†}
%%%\fnsymbol{footnote}

\begin{table}[H]
	\begin{center}
		\caption{Influence of the edge detection operator. 
        % The UWSegFormer in this experiment only incorporates UIQA and MAA.
        }
		\label{table:edo}
		\begin{tabular}{c|c|c} % <-- Alignments: 1st column left, 2nd middle and 3rd right, with vertical lines in between
			\hline
			Model & Edge detection operators  &  mIoU↑\\
			\hline
			UWSegFormer & Laplacian operator &  81.68 \\
			UWSegFormer & Robert operator &  81.70 \\
			UWSegFormer & Prewitt operator & 81.76 \\
			UWSegFormer & Sobel operator & 82.01\\
			UWSegFormer & Scharr operator & \textbf{82.12}\\
			\hline
		\end{tabular}
	\end{center}
\end{table}\par
	\subsection{MAA Generalizability Study}
	%Through the ablation experiments, we conclude that MAA can effectively reduce the computational burden while still maintaining stable performance. 
	To assess the generalization performance of MAA, we test it with two different backbones (the Swin and LVT), as shown in Tab.~\ref{table:4}. It is evident that Swin+MAA improves the performance by 0.2 in terms of mIoU, while reducing the number of parameters by almost half, from 58.94M to 29.27M, and significantly decreasing GFlops, thereby reducing the computational burden by 242.95G. 
	For the lightweight LVT model, MAA can also reduce the parameters by 2.4M and almost halve the computational burden. \par
	The results demonstrate the versatility and effectiveness of MAA, which allows models with a four-layer feature extraction structure to be applied to underwater environments while effectively reducing the number of parameters and computational burden.
	
	\setcounter{footnote}{0}
	\renewcommand{\thefootnote}{*}
	%%\fnsymbol{footnote}
	
	\begin{table}[H]
		\begin{center}
			\caption{MAA generalizability testing of Swin and LVT models. Methods with
				* use MAA.}
			\label{table:4}
			\begin{tabular}{c|c|c|c} % <-- Alignments: 1st column left, 2nd middle and 3rd right, with vertical lines in between
				\hline
				Models & Params(M)↓  & GFlops↓ & mIoU↑\\
				\hline
				Swin\cite{RF11} & 58.94 & 276 & 80.70 \\
				Swin\footnote{Swin, symbol.} & \textbf{29.27} & \textbf{33.05} & \textbf{80.90} \\
				\hline
				LVT\cite{RF32} & 3.83 & 10.51 & 80.72\\
				LVT\footnote{LVT, symbol.} & \textbf{3.59} & \textbf{5.29} & \textbf{80.99}\\
				\hline
			\end{tabular}
		\end{center}
	\end{table}\par
	\begin{table}[H]
		\begin{center}
			\caption{Comparison with SOTA methods on SUIM and DUT. The bolded experimentation is our method}
			\label{table:5}
			\begin{tabular}{c|c|c|c|c|c|c} % <-- Alignments: 1st column left, 2nd middle and 3rd right, with vertical lines in between
				\hline
				Method & Encoder  & GFlops↓  & \multicolumn{2}{c|}{SUIM} & \multicolumn{2}{c}{DUT}\\
				\cline{4-7}
				&  &  & FPS↑ & mIoU↑ & FPS↑ & mIoU↑ \\
				\hline
				FCN\cite{RF25}&	MobileNetV2&	46.11 &	55.75 &	66.40&	54.45&	62.75\\
				UNet\cite{RF41}&	Resnet50&	238 &	20.55 &	56.12&	18.46&	59.15\\
				PSPNet\cite{RF42}&	MobileNetV2&	61.68 &	59.43 &	69.76&	59.40& 67.84\\
				DeepLabV3+\cite{RF23}&	MobileNetV2&	80.06 &	49.16 &71.03&	47.03&	66.68\\
				SeaFormer\cite{RF43}&	SeaFormer-L&	7.55 &	64.04 &	65.23&	61.42&	57.54\\
				PIDNet\cite{RF44}&	PIDNet-s&	6.96 &	86.95 &	71.46&	92.45&	67.79\\
				DDRNet\cite{RF45}&	DDRNet-s&	5.35 &79.67 &	73.44&	74.51& 69.15\\

                USS-NET\cite{RF49}&	Resnet50&	342 & 16.14 &	72.09&	15.23& 67.87\\
				Swin\cite{RF11}&	Swin-Tiny&	276 &	16.32  &	80.70&	17.09&	69.35\\
				SegFormer\cite{RF13}&	MiT-B0&	7.94 &	\textbf{95.97} &	80.57&	\textbf{95.21}&	70.81\\
                % Improved SegFormer\cite{RF48}&	Swin-Tiny&	26.51 &	35.24  & 77.00&	33.67&	67.21\\
				\hline
				UWSegFormer&	MiT-B0+UIQA&	\textbf{3.21} &	62.42 &	\textbf{82.12}&	62.25&	\textbf{71.41}\\
				\hline
				%			Swin[11] & 58.94 & 276 & 80.70 \\
				%			Swin+MAA & 29.27 & 33.05 & 80.90 \\
				%			LVT[32] & 3.83 & 10.51 & 80.19\\
				%			LVT+MAA & 3.59 & 5.29 & 80.99\\
			\end{tabular}
		\end{center}
		\label{tabel:5}
	\end{table}\par
	\subsection{Comparison to SOTA methods}
	Tab.~\ref{table:5} summrizes the results for GFlops, latency and mIoU on the SUIM and DUT datasets. As shown in Tab.~\ref{table:5}, UWSegFormer achieves 82.12$\%$ mIoU using 3.21 GFlops on the SUIM dataset, outperforming all other semantic segmentation methods at a lower computational burden while still obtaining a higher mIoU. For instance, it achieves a frame rate of 46.1 FPS, outperforming Swin (Swin-Tiny)\cite{RF11} while simultaneously enhancing mIoU performance by 1.42$\%$. When compared to the latest methods such as PIDNet\cite{RF44} and DDRNet\cite{RF45}, our method shows an improvement of 10.66$\%$ and 8.68$\%$ in mIoU, respectively.
	
	In addition, our method can be applied to the DUT dataset. UWSegFormer has a lower computational burden of 4.73 GFlops compared to SegFormer-B0 and increases mIoU by 0.6$\%$. %Compared to CNN-based methods, our method has an absolute advantage. 
	UWSegFormer achieves a frame rate of 62.25 FPS and  a mIoU of 71.41$\%$, representing an improvement of approximately 4.73$\%$ in mIoU and 15.22 FPS in speed compared to DeeplabV3+. 
	\begin{figure}[H]
		\centering
		\includegraphics[width=1\linewidth]{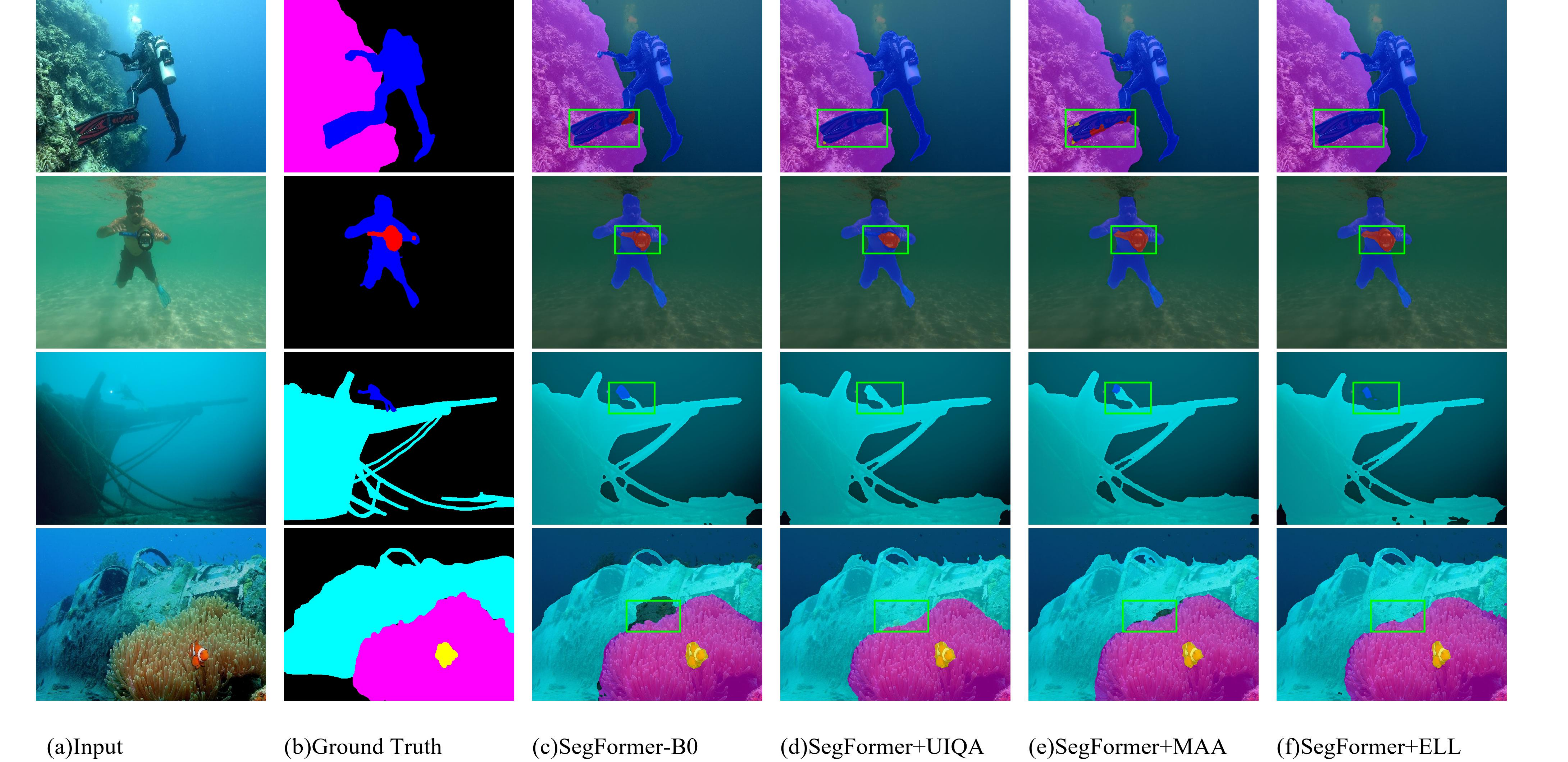}
		\caption{Qualitative results on SUIM validation set. (a) Input images. (b) Ground truth. (c) B0 as backbone SegFormer. (d) Only added UIQA. (e) Only added MAA. (f) Only added ELL. Green parts emphasize the enhancement in segmentation results by various modules.}
		\label{fig:6}
	\end{figure}
	\begin{figurehere}
		\centering
		\includegraphics[width=0.99\linewidth]{figures/combined_photos2_3}
		\caption{Qualitative segmentation results on the SUIM and DUT validation val sets. (a) Input images. (b) GT. (c) FCN. (d) PSPNet. (e) DeepLabV3+. (f) Unet. (g) SeaFormer. (h) PIDNet. (i) DDRNet. (j) Swin. (k) SegFormer-b0. (l) UWSegFormer (ours).}
		\label{fig:7}
	\end{figurehere}
	\subsection{Qualitative Results.} \par
	We visualise the improvement in model performance for each component. In Fig~\ref{fig:6}(d) demonstrates a significant improvement in performance compared to Fig.~\ref{fig:6}(c), where the model is concentrated on high-quality image channels, resulting in more accurate model predictions. In comparison to Fig.~\ref{fig:6}(c), Fig.~\ref{fig:6}(e) demonstrates that the model captures the detailed information in the underwater image. Meanwhile, Fig.~\ref{fig:6}(f) shows that the model can be guided to obtain a more complete boundary when ELL is introduced.\par
	Fig.~\ref{fig:7} displays the results of different methods on the SUIM and DUT validation sets. UWSegFormer outperforms other methods with the assistance of the UIQA, MAA and ELL. Our approach demonstrates higher accuracy in predicting pixel labels in underwater environments. The results indicate that our methodology excels in handling challenging scenarios, particularly demonstrating significantly improved performance in ambiguous regions.
	
	\section{Conclusion}
	This study presents UWSegFormer, a semantic segmentation approach for underwater environments. We focuse on high-quality semantic information channels through the UIQA module to enhance feature representation, thereby meeting segmentation requirements in low illumination environments. Furthermore, the innovative MAA module concentrates on detail information in underwater images by efficiently aggregating features. Concurrently, the incorporation of ELL enables the model to focus on crucial boundary information. Experimental results conducted on SUIM and DUT datasets achieved 82.12 $\%$ and 71.41 $\%$ mIoU, respectively, demonstrating the effectiveness of the proposed UWSegFormer in underwater environments.
	% \noindent \textbf{Limitation.} 
    The UIQA and MAA introduced in this study utilize a four-layer feature extraction architecture similar to SegFormer. Our aim is to increase the modules' adaptability by allowing them to be adjusted to other architectures in future work.  \par
	\textbf{Acknowledgement:} This work was supported in part by NSF of China under Grant No. 61903164 and in part by NSF of Jiangsu Province in China under Grants BK20191427.
	%% The Appendices part is started with the command \appendix;
	%% appendix sections are then done as normal sections
	%% \appendix
	
	%% \section{}
	%% \label{}
	
	%% If you have bibdatabase file and want bibtex to generate the
	%% bibitems, please use
	%%
	%%  \bibliographystyle{elsarticle-num} 
	%%  \bibliography{<your bibdatabase>}
	
	%% else use the following coding to input the bibitems directly in the
	%% TeX file.
	\bibliographystyle{elsarticle-num}
	\bibliography{IEEEexample}
	%\begin{thebibliography}{00}
	%
	%%% \bibitem{label}
	%%% Text of bibliographic item
	%
	%\bibitem{RF1}1111
	%
	%\end{thebibliography}
	
\end{document}